\documentclass{article}

\usepackage{arxiv}
\usepackage{eurosym}
\usepackage[utf8]{inputenc} 
\usepackage[T1]{fontenc}    
\usepackage{url}            
\usepackage{booktabs}       
\usepackage{amsfonts}       
\usepackage{nicefrac}       
\usepackage{microtype}      
\usepackage{lipsum}
\usepackage{graphicx}
\usepackage{color}
\usepackage{amssymb}
\usepackage{subcaption}
\usepackage{amsmath}

\title{Appliance-Level Monitoring with Micro-Moment Smart Plugs}

\author{
  Abdullah Alsalemi\thanks{This paper has been accepted in SCA2020: The Fifth international conference on Smart City Applications}, Yassine Himeur, Faycal Bensaali\\
  Department of Electrical Engineering\\
  Qatar University\\
  Doha, Qatar \\
  \texttt{yassine.himeur@qu.edu.qa;a.alsalemi@qu.edu.qa;f.bensaali@qu.edu.qa} \\
   \And
 Abbes Amira \\
  Institute of Artificial Intelligence\\
  De Montfort University\\
  Leicester, United Kingdom \\
  \texttt{abbes.amira@dmu.ac.uk} \\
}

\begin{document}
\maketitle

\begin{abstract}
Human population are striving against energy-related issues that not only affects society and the development of the world, but also causes global warming. A variety of broad approaches have been developed by both industry and the research community. However, there is an ever increasing need for comprehensive, end-to-end solutions aimed at transforming human behavior rather than device metrics and benchmarks. In this paper, a micro-moment-based smart plug system is proposed as part of a larger multi-appliance energy efficiency program. The smart plug, which includes two sub-units: the power consumption unit and environmental monitoring unit collect energy consumption of appliances along with contextual information, such as temperature, humidity, luminosity and room occupancy respectively. The plug also allows home automation capability. With the accompanying mobile application, end-users can visualize energy consumption data along with ambient environmental information. Current implementation results show that the proposed system delivers cost-effective deployment while maintaining adequate computation and wireless performance.
\end{abstract}


\section{Introduction} \label{sec1}
Undoubtedly, energy saving and energy security have become major contemporary issues. We are facing an energy shortage that not only affects the world's economy, environment, and growth, but also results in global warming \cite{dileep2020survey}. A set of recent developments is about to change this picture and propose effective energy policies. These policies explicitly or indirectly create drivers that, from the point of view of business and end-users, are perceived to be any activity taken by the energy-efficient system created. In Qatar, the goal of the TARSHEED initiative is to raise awareness of energy-saving activities and the unnecessary energy use in the country as a whole \cite{kaabi2012conservation}. This has been achieved across a variety of advertising promotions, standards, and competitions \cite{tarhseed_qatar_2020}. 

Smart metering is instrumental in collecting and analyzing energy market data \cite{WOUTERS201522}. The smart meter is an important tool for managing the energy consumption curve efficiently. This calls for a connection with the quantity of usage and the quantity of production, allowing for the substitution of flat-rate prices with better strategies \cite{osareti_2016_smart}. There is, therefore, a need for development and standardization of metering schemes. 

One of the main elements of smart metering is tracking the volume of the system. This enables processing of unique data linked to each device without the need for aggregation algorithms. This is why smart plugs can play a significant role in the monitoring and operation of domestic appliances, where end-users mount them for each device at a fairly low cost, and thus, obtaining the benefit of precise calculation of power usage and home automation. 

Throughout this sense, we introduce micro-moments, which are time-based time-slots where the end-user uses an appliance or occupy a space (e.g. room, corridor, hall, etc.) \cite{alsalemi2019ieeesystems}. This concept enables the extraction of different moments when the end-user engages in unhealthy energy consumption patterns, making it easier to classify and produce recommendations to enhance the energy efficiency of the household. 

Throughout this article, we are presenting a micro-moment smart plug as part of the broader $(EM)^3$ project, which seeks to use artificial intelligence to achieve high domestic energy performance. The $(EM)^3$ smart plug enables real-time power measurements to be collected in addition to environmental information, such as temperature , humidity, and occupancy. 

The remainder of this paper is structured as follows. Section \ref{sec2} outlines the latest research on smart plugs. Section \ref{sec3} provides a description of the broader $(EM)^3$ structure and its elements. The smart plug is described in Section \ref{sec4}, and data visualization scenarios are described in Section \ref{sec5}. System implementation results are expounded upon in Section \ref{sec6}. The paper is concluded in Section \ref{sec7}.

\section{Related Work}
\label{sec2}

In this section, an overview of recent developments to smart plugs is provided. First, Ahmed et al. proposed a smart plug prototype that measures power consumption in home energy management systems \cite{ahmed_smart_2015}. It is assisted by a Zigbee microcontroller. It has been demonstrated, from their findings, that the suggested plug absorbs less power and achieves better precision in contrast with the oscilloscope. In fact, the system provides the option to connect/disconnect the attached device from the power supply. 

In addition, Hajahan and Anand have suggested an Arduino microcontroller-based smart plug that uses ENC28J60 for communication \cite{altaf_hamed_shajahan_data_2013}. This is powered by a split core style current transformer for non-invasive current calculation and an Android-based user interface. 

To alleviate the problems of peak shortages, Ganu et al. have suggested a cost-effective smart plug that monitors the synchronization (by flipping on or off) of loads to the grid at on or off on times \cite{ganu_nplug_2012}. This is done with the help of real-time analysis and data collection \cite{arjunan_2015}. It allows off-peak and unpredictable scheduling by addressing user choice settings in addition to grid load conditions. 

Another research relevant to in-device detection, Petrovic et al. suggested a smart plug for electrical load detection focused on the \cite{petrovic_active_2017} active sensing method. The input signal produced by the smart plug is modified before the output is calculated in such a way that more distinct real-time data is created. An artificial neural network is used to evaluate the approach to specify classification performance metrics; accuracy and speed. 

Environmental control is another important aspect. That is why Gomes et al. have suggested a smart plug that is mindful of the environmental factors and awareness of the device resource background \cite{gomes_intelligent_2018}. The technique adopts a multi-agent structure strategy that helps the agent to react to any adjustments that arise and to communicate with other agents. 

To make progress on the literature studied, the main contributions of this paper are summarized as follows: 

\begin{itemize} 
\item Modern application of the concept of micro-moments in the control of energy use. 
\item Design and implementation of a micro-moment based smart plug for collecting contextual details, such as temperature, humidity, brightness and room occupancy. 
\item The smart plug enables edge computing features, such as the identification of several devices attached to the smart plug. 
\end{itemize}
\section{The EM3 Energy Efficiency Framework}  \label{sec3}
The Consumer Engagement Towards Energy Saving Behavior by means of Exploiting Micro Moments and Mobile Recommendation Systems $(EM)^3$ platform has been developed to promote customer behavioral improvement through increasing energy use understanding \cite{alsalemi2019ieeesystems}. 
The $(EM)^{3}$ system consists of the following components \cite{alsalemi_access_2020}: 

\begin{enumerate}
\item Data Collection: gathers data dependent on micro-moments for power usage and environmental monitoring \cite{alsalemi_classifier_2019}. 
\item Classification: detects and analyzes abnormal energy patterns \cite{HIMEUR_2020}.
\item Suggestions and Automations: provides personalized guidance to end-users to endorse energy saving management activities coupled with recommended actions for environmental change \cite{sardianos_smartgreens_2019} .
\item Visualization: upload results, observations and feedback in an accessible and engaging manner via a mobile application. 
\end{enumerate} 

Sensing devices play an essential role in capturing data in a given datastore \cite{alsalemi_rtdpcc_2019}. They are used for wirelessly uploading collected data to a testing center at Qatar University (QU). The study lab consists of a variety of testing cubicles, some of which are installed with sensing instruments to the $(EM)^3$ storage server housed in the QU building. Each cubicle contains data aggregated from monitors, screens and table lamps. 

The No-SQL CouchDB cloud platform is used to store customer micro-moments and usage levels, user expectations and resources, energy management guidelines and ranking ratings. 
The dataset used in this study, Qatar University Dataset (QUD), is compiled in a micro-moment laboratory at QU \cite{alsalemi_rtdpcc_2019}. The set-up collects environmental sensors (indoor temperature and humidity, room luminosity and motion sensors) and power consumption for a number of appliances (i.e. light bulb, computer). Data is collected using wireless sensing modules that communicate in real-time to the backend. QUD contains data points divided by a couple of seconds each. The dataset was collected earlier before this work.

\section{Proposed System Design}
\label{sec4}

The $(EM)^3$ smart plug helps to combine both energy tracking at the device level and the environmental details of the current household space. Data on the energy usage of equipment is assumed to be the key piece of knowledge, whereas environmental factors such as temperature, humidity, luminosity and occupancy provide a background under which the knowledge on use is evaluated. The system is composed of two sub-units: the power consumption unit and the environmental monitoring unit.

\begin{figure*}[!ht]
\centering
\includegraphics[trim={0in 0in 0in 0in},clip,width=1\linewidth]{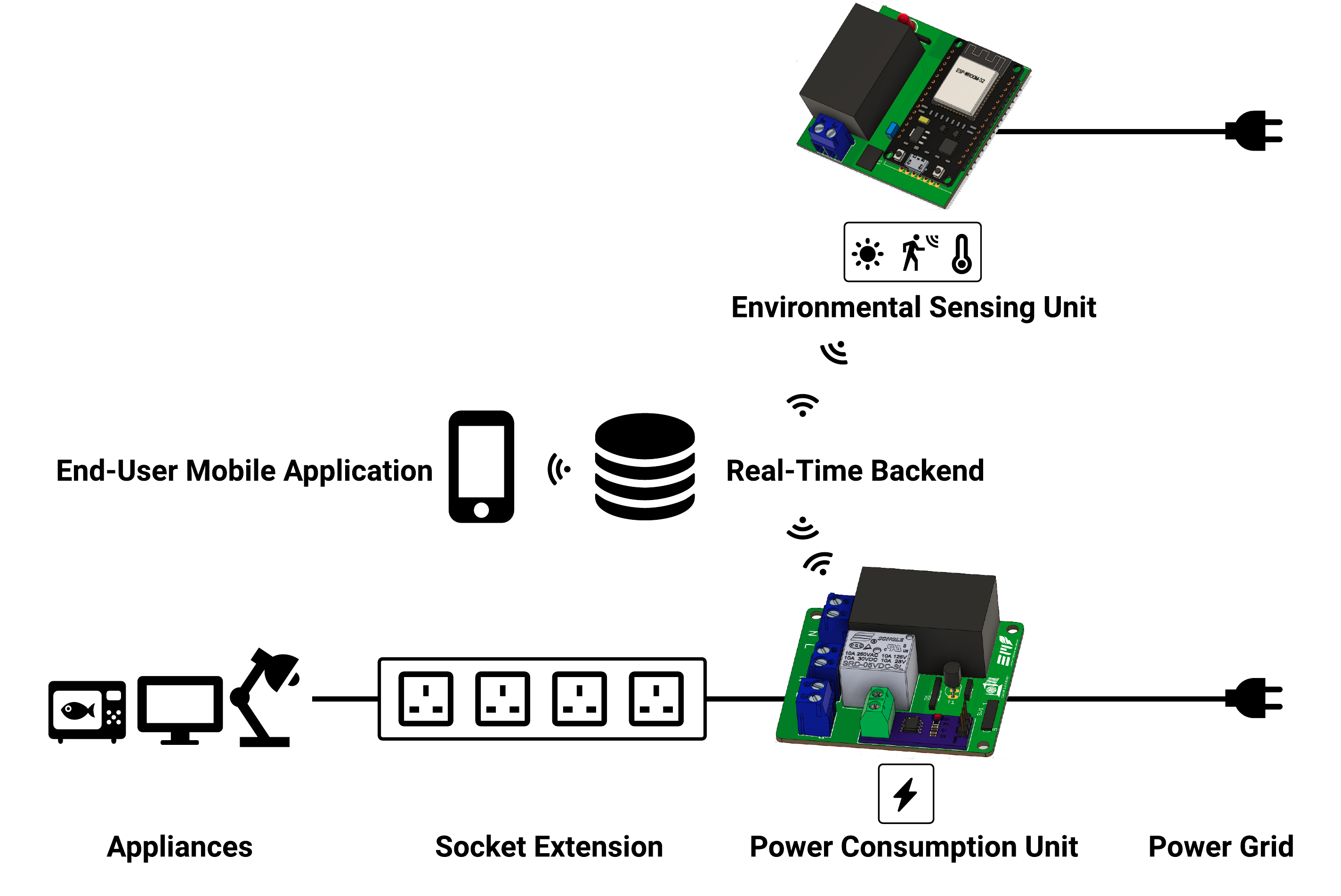}
\caption{Overview of the micro-moment smart plug.}
\label{fig:smart-plug-block} 
\end{figure*}

\subsection{Power Consumption Unit}

Fig. \ref{fig:smart-plug-block} provides a description of the $(EM)^3$ smart plug. The power wire from the grid travels into the smart plug and goes to the extension socket where the appliance(s) can be worked. Smart plug modules calculate the current used by the equipment attached to the extension cord and compare it with the normal voltage of the nation in which it is worked. The HLW8012 intrusive power sensor was used to calculate current values up to 20A with 5\% tolerance.

\subsection{Environmental Monitoring Unit}

The environmental unit is used to measure temperature, humidity, luminosity and occupancy of a given household's room. The data is transmitted in real-time to the backend for further processing.

The HC-SR501 motion sensor is used to assess space occupancy and the DHT-22 temperature and humidity sensor operates between -40-80$^{\circ}$C and 0-100\% for ambient temperature and relative humidity, respectively. Also, the light sensor TSL2591, which can sense light in the range of 0.1-40,000 Lux. The sensors are described in Table \ref{tab:modules-table}. 

In addition, real-time synchronization between the chosen micro-controller and the server side is accomplished where a delay of only a millisecond is required. Real-time synchronization is checked by matching the time stamp on the sensor-side microcontroller with the time stamp on the server-side.

Data obtained from various data collection programs were stored on the server side. Data fusion algorithms are used to obtain data and provide a full overview of the world \cite{himeur_fusion_2020}. In fact, it provides scalability if further sensors are to be installed. Instead, machine learning classifiers are educated on cumulative data where they can be used to grasp how energy use looks like \cite{himeur_2020_2}. The efficiency of such classifiers can be checked by measuring precision, memory, responsiveness, F1 score and accuracy, along with a confusion matrix \cite{himeur_2020_3}. The recommendation framework \cite{sardianos_rehab_2020}, focused on these algorithms, would communicate advice and recommendations persuading customers of energy from their smartphones \cite{sardianos2020data}. Once excessive energy consumption has been detected, suggestions are sent to users to advise them on how to reduce it. The simulation of energy consumption data, the use of existing datasets and the setting up of the actual environment are all used to check the validity of the recommendations provided. 

In addition, since the smart plug can be connected to a socket extension, multiple appliances can be connected and identified. It can inhibit the accuracy of understanding what device absorbs how much electricity. For this purpose, we are focusing on the implementation of a range of appliance recognition algorithms to recognize the appliances used with high precision \cite{HIMEUR_2020}. It is noteworthy to mention that the internal power consumption of the plug is designed to minimally affect the overall consumption and multi-appliance identification is under-development.

\begin{table}[ht]
    \caption{Smart plug sensors}
    \label{tab:modules-table}
    \centering

\begin{tabular}{| p{0.2\textwidth}| p{0.5\textwidth}| p{0.3\textwidth} |}
\hline
\textbf{Name} & \textbf{Description} & \textbf{Components Used} \\
\hline
Energy monitoring & Measures appliances power consumption in Watts. & ACS712 invasive Hall effect current sensor \\ 

Occupancy & Detects whether room is occupied. Selected for accuracy. & AM312 motion sensor \\ 

Temperature and humidity & Measures indoor ambient temperature. Selected for cost-efficiency and adequate accuracy. and relative humidity & DHT22 \\ 

Luminosity & Measures room's luminosity in Lux. Selected for accuracy and wide range. & Adafruit TSL2591 \\ 

\hline
\end{tabular}

\end{table}

\section{Consumer Data Visualization}
\label{sec5}

\begin{figure*}[!ht]
\centering
\includegraphics[trim={0in 0in 0in 0in},clip,width=0.7\linewidth]{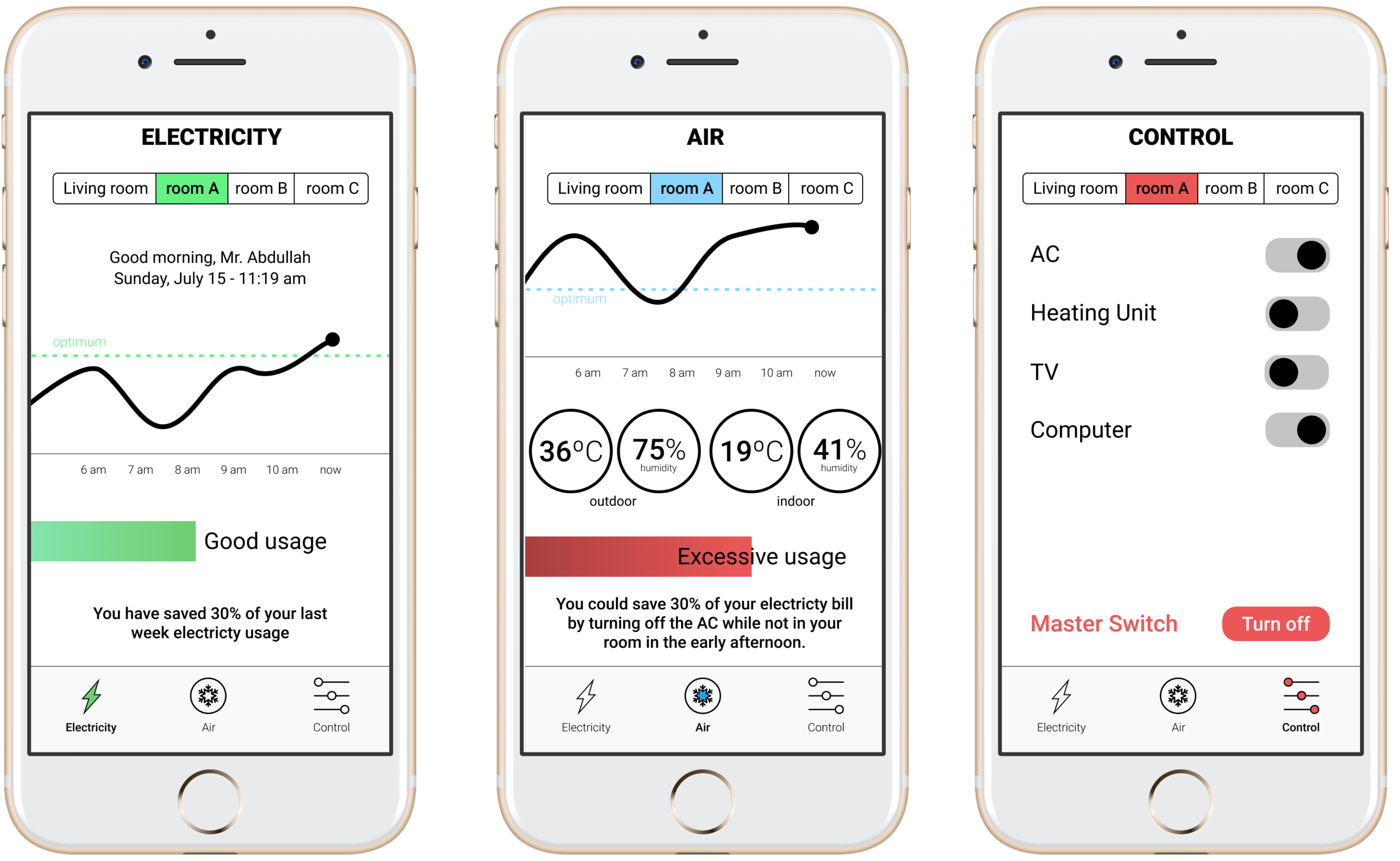}
\caption{The $(EM)^3$ mobile application: (a) the electric power consumption, (b) air-related consumption, and (c) appliance control screens.}
\label{fig:app} 
\end{figure*}

After data collection and preliminary data analysis, data visualization is provided through a mobile application where users can actually see their consumption. In addition, meaningful information and data are provided to moderate users' behavior towards energy efficiency. In this way, the smart plug data is easily displayed and the signals behind it are shown. 

The $(EM)^3$ smartphone device seeks to encourage energy conservation by visualizing smart plug data in real-time. Fig. \ref{fig:app} shows the main screens of the first version of the $(EM)^3$ mobile app. Line plots are commonly used to provide data along with a tiny overview that offers supplementary detail on the chosen device (as seen in Fig. \ref{fig:app}a and \ref{fig:app}b). Additional energy efficiency standards are shown in \ref{fig:app}a and ambient environmental information (i.e. indoor and outdoor temperature and humidity, indoor illumination and room occupancy). 

A data visualization analysis has been undertaken to evaluate the right visualization for energy end-users in order to improve the field of data visualization. Best visualizations for the current usage of energy data visualization are suggested\footnote{http://em3.qu.edu.qa/index.php/data-visualization-app}.

\section{Results}
\label{sec6}

This segment outlines the latest implementation and performance of the new smart micro-moment connection. In terms of configuration, the smart plug comprises a printed circuit board (PCB), a 3D-printed plastic casing, a connector and an extension of the outlet. The results are described for both system's sub-units.

\subsection{Power Consumption Unit} 
As the current smart micro-moment plug requires several devices to be linked at the same time, it is of vital importance to incorporate an appliance identification program that can recognize each unit utilizing its power consumption signature. In this regard, we are presenting a simple but successful method focused on first detecting appliance events using a cepstrum-dependent detector defined in \cite{HIMEUR_2020}. Following, a combination of two time-domain attribute extraction algorithms, including root mean square (RMS) and mean absolute deviation (MAD) is used to identify each linked system utilizing observed events. 

In this way, various machine learning algorithms are deployed to identify five types of appliances utilizing specific parameter settings, namely support vector machine (SVM), K-nearest networks (KNN), decision tree (DT), deep neural networks (DNN) and decision bagging tree (DBT). In fact, a 10 USD cross-validation contract has been used to test the existing device recognition method. Table \ref{CombPerf} shows the accuracy and the results of the F1 score obtained using the proposed summation based fusion technique compared to the use of the RMS and MAD descriptors, separately. It is clear that this fusion strategy can improve identification accuracy by 2.46\% and 2.38\% compared to MAD and RMS, respectively. In comparison, the F1 value was improved by 3.34\% and 2.59\%, respectively, relative to MAD and RMS. It is worth mentioning that smart plug device recognition capabilities are in the early stages of growth.

\begin{figure}[!ht]
\centering
\includegraphics[trim={0in 0in 0in 0in},clip,width=0.6\linewidth]{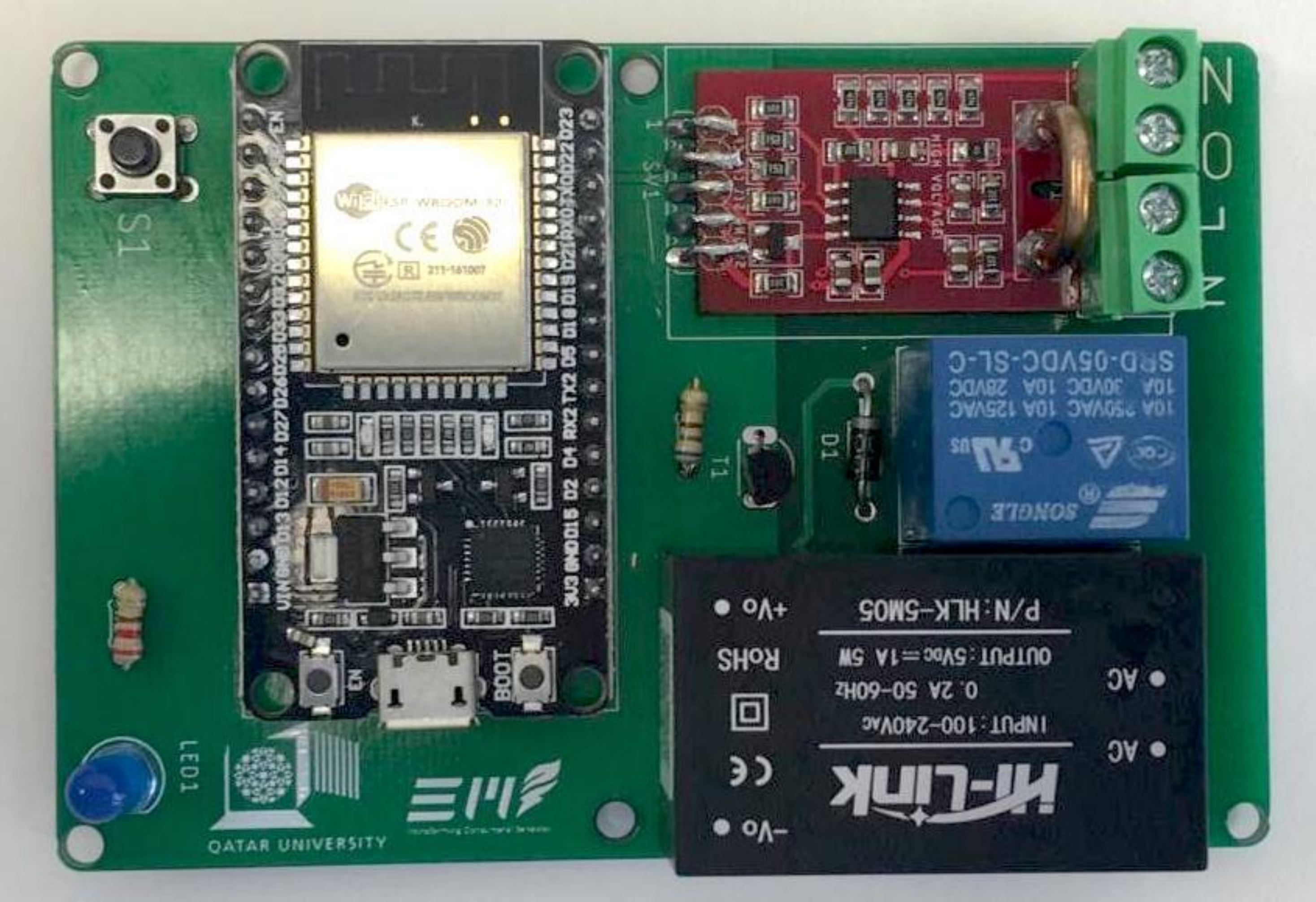}
\caption{Power consumption unit PCB.}
\label{fig:pcb} 
\end{figure}

\begin{table*}[t!]
\caption{Performance of the proposed descriptor fusion used to recognize electrical appliances.}
\label{CombPerf}
\begin{center}

\begin{tabular}{lccccccc}
\hline
{\small \textbf{ML} } & {\small \textbf{Classifier}} & \multicolumn{2}{c}{ \ 
{\small \textbf{RMS \ }}} & \multicolumn{2}{c}{{\small \textbf{MAD}}} & 
\multicolumn{2}{c}{{\small \textbf{Fusion}}} \\ \cline{3-8}\cline{3-8}
{\small \textbf{algo}} & {\small \ \textbf{parameters}} &  \ \ \ \ {\small 
\textbf{Acc\ \ \ \ \ }} &  \ \ \ \ \ {\small \textbf{F1\ \ \ \ \ }} &  \ \ \
\ {\small \textbf{Acc \ \ \ }} &  \ \ \ \ \ {\small \textbf{F1 \ \ \ \ }} & 
\ \ \ \ {\small \textbf{Acc\ \ \ \ \ }} &  \ \ \ \ \ {\small \textbf{F1 \ \
\ \ \ }} \\ \hline
{\small SVM } & {\small Linear Kernel} & {\small 89.05} & {\small 88.74} & 
{\small 88.22} & {\small 88.74} & {\small 91.76} & {\small 91.59} \\ 
{\small SVM} & {\small Quadratic kernel} & {\small 90.22} & {\small 89.57} & 
{\small 89.22} & {\small 88.57} & {\small 92.83} & {\small 91.57} \\ 
{\small SVM} & {\small Gaussian kernel} & {\small 91.63} & {\small 90.6} & 
{\small 91.27} & {\small 89.71} & {\small 92.7} & {\small 92.23} \\ 
{\small KNN} & {\small K=1/Euclidean distance } & {\small 92.24} & {\small %
91.19} & {\small 93.24} & {\small 90.19} & {\small 93.95} & {\small 93.66}
\\ 
{\small KNN \ \ } &   {\small K=10/Weighted Euclidean dist \  } & {\small %
93.65} & {\small 92.33} & {\small 93.11} & {\small 91.96} & {\small 94.92} & 
{\small 94.85} \\ 
{\small KNN} & {\small K=10/Cosine dist} & {\small 90.75} & {\small 89.43} & 
{\small 89.75} & {\small 89.43} & {\small 92.87} & {\small 92.7} \\ 
{\small DT} & {\small Fine, 100 splits} & {\small 93.94} & {\small 93.69} & 
{\small 93.59} & {\small 93.22} & {\small 95.63} & {\small 95.51} \\ 
{\small DT} & {\small Medium, 20 splits} & {\small 90.79} & {\small 90.17} & 
{\small 90.44} & {\small 90.32} & {\small 93.44} & {\small 93.11} \\ 
{\small DT} & {\small Coarse, 4 splits} & {\small 87.92} & {\small 86.83} & 
{\small 87.77} & {\small 87.4} & {\small 90.57} & {\small 90.41} \\ 
{\small DNN} & {\small 50 hidden layers} & {\small 93.25} & {\small 92.68} & 
{\small 92.22} & {\small 91.55} & {\small 95.11} & {\small 94.78} \\ 
{\small \textbf{DBT}} & {\small \textbf{30} \textbf{learners, 42 k splits}}
& {\small \textbf{96.41}} & {\small \textbf{95.93}} & {\small \textbf{96.33}}
& {\small \textbf{95.18}} & {\small \textbf{98.49}} & {\small \textbf{98.32}}
\\ \hline
\end{tabular}

\end{center}
\end{table*}

In addition, the impact of the fusion strategy on the performance of the proposed recognition of appliances has been assessed. The findings indicate strong performance rates obtained by combining TD descriptors in respect to the DBT classifier (30 learners, 42 k splits) and considering three fusion strategies. It is obvious that the summation-based approach produces the best precision and F1 ratings. Specifically, 98.79\% accuracy and 98.52\% F-score were achieved by combining the number, while 96.98\% accuracy and 96.49\% F1 score and 97.43\% accuracy and 97.11\% F1 score were achieved by combining and multiplication approaches, respectively.

In terms of hardware, the PCB is the heart of the power plug. The board, seen in the Fig. \ref{fig:pcb}, features a self-powered system, eliminating the need for a specific power source to run it. In addition, a relay is installed to allow remote control of the device, and intrusive energy monitoring is used due to direct contact to the device. 

The PCB is designed to handle two types of micro-controllers, the most fitting of which are assisted by two micro-controllers. Both the Arduino MKR-1010 and the ESP32 can be assisted. In fact, the (EM)$^3$ smart plug is projected to greatly accelerate the implementation of domestic energy use control systems worldwide. This can be manufactured at a cost of between 20 USD and 40 USD. Performance-wise, the micro-controllers used will do fairly sophisticated computation in addition to real-time wireless connectivity.

\subsection{Environmental Monitoring Unit}

\begin{figure}[!ht]
\centering
\includegraphics[trim={0in 0in 0in 0in},clip,width=0.6\linewidth]{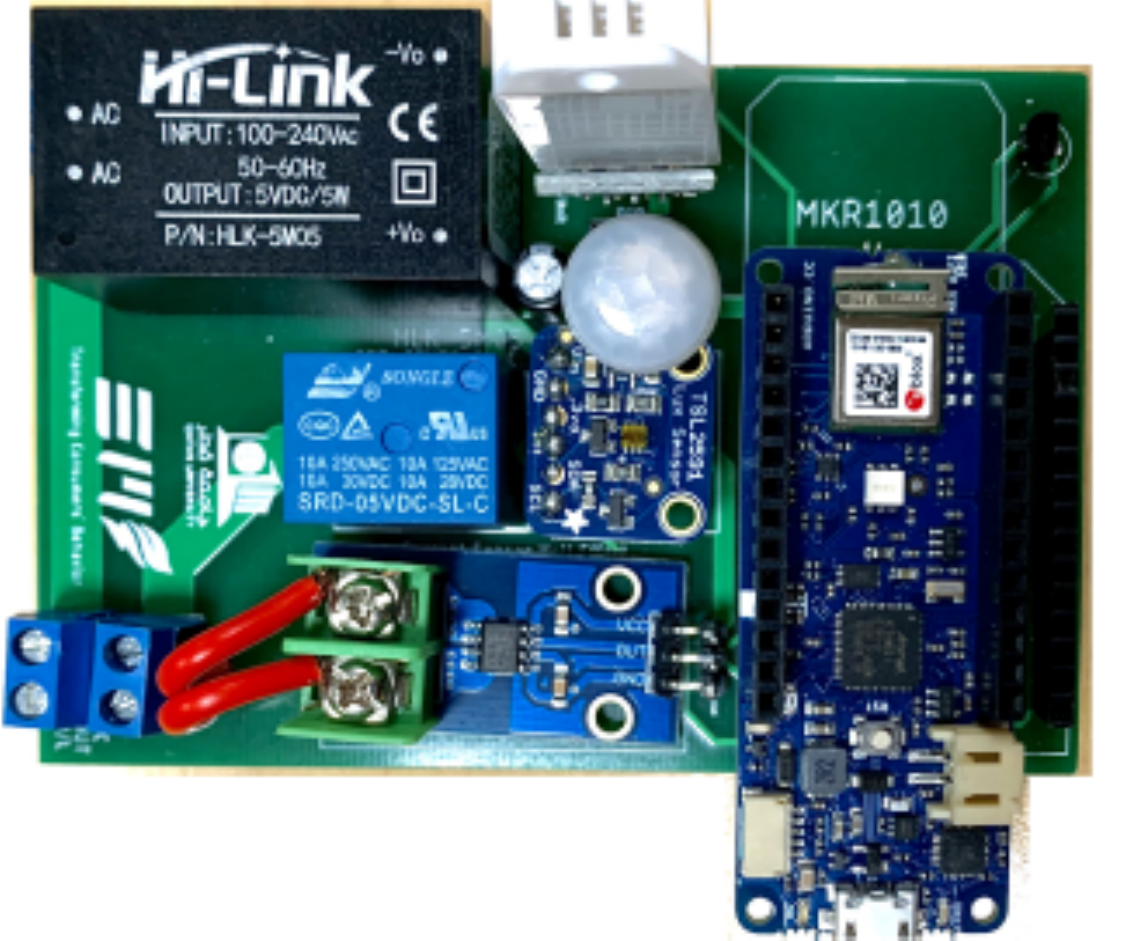}
\caption{Environmental sensing unit PCB.}
\label{fig:pcb2} 
\end{figure}

Similarly to the power consumption unit, the environmental monitoring unit is housed in a PCB that contains the micro-controller and the sensors needed to monitor temperature, humidity, luminosity, and presence. The micro-controller is an ESP32, which supports wireless transmission to the CouchDB backend after preliminary post-processing. In effect, on-board processing reduces noise in data as well as the packet size. Performance, communication latency, and costs have been computed and compared in Table \ref{tab:results-table}. The smart plug micro-controller has been programmed with a platform-agnostic Arduino program. The PCB is shown in Fig. \ref{fig:pcb2}.

\begin{table}[ht]
    \caption{Smart plug performance per micro-controller}
    \label{tab:results-table}
    \centering
		
\begin{tabular}{ | p{0.2\textwidth} | p{0.2\textwidth}| p{0.3\textwidth} | p{0.15\textwidth} |}
\hline
\textbf{Used Board Name} & \textbf{Processing Speed (s)} & \textbf{Communications Latency (s)} & \textbf{Cost (USD)} \\
\hline
ESP-WROOM-32 & 0.16 & 3.19 & 10 \\ 

Arduino MKR 1010 & 1.05 & 2.25 & 33.90 \\ 

\hline
\end{tabular}

\end{table}

The concept of a smart plug that processes data before pushing it to the cloud is linked to the trend of edge computing, where some processing takes place where data is produced, saving communication resources and increasing efficiency. 

Compared to the literature reviewed, the proposed approach provides the advantage of micro-moment extraction, which allows for a more accurate analysis of daily consumption. It also enables several devices to be attached at the same time.



However, there is indeed a range of drawbacks of the new application. The thickness of the device is a downside that is deemed cumbersome relative to many current solutions. In addition, in terms of cyber security, the current implementation lacks the various cyber-attack defense mechanisms that will be considered in future publications. Eventually, a more computationally efficient board such as the ESP32-S2 may be used to help operate more complicated in-chip classification algorithms.

\section{Conclusions}
\label{sec7}

In this paper, a micro-moment smart plug is proposed as part of the $(EM)^3$ framework. The smart plug, which includes two sub-units: the power consumption unit and environmental monitoring unit collects energy consumption of appliances along with contextual information such as temperature, humidity, luminosity, and room occupancy respectively. The plug also allows home automation capabilities. With the $(EM)^3$ mobile app, end-users can see visualized power consumption data along with ambient environmental information. In addition, the foundations of the appliance recognition method was successfully verified by combining two time-domain descriptors, resulting in high accuracy and F1 score benchmarks.

\section*{Acknowledgments}
This paper is made possible by National Priorities Research Program (NPRP) grant No. 10-0130-170288 from the Qatar National Research Fund (a member of Qatar Foundation). The statements made herein are solely the responsibility of the authors.

\end{document}